\documentclass[10pt,twocolumn,letterpaper]{article}

\PassOptionsToPackage{table}{xcolor}
\usepackage[pagenumbers]{cvpr} 
\usepackage[table]{xcolor}

\usepackage{lineno}

\definecolor{cvprblue}{rgb}{0.21,0.49,0.74}
\usepackage[pagebackref,breaklinks,colorlinks,allcolors=cvprblue]{hyperref}
\usepackage{algorithm}
\usepackage{algorithmic}
\usepackage{amsmath}
\usepackage[capitalize]{cleveref}
\usepackage{algorithm}
\usepackage{multicol}
\usepackage{multirow}
\usepackage{makecell}
\usepackage{pifont}
\usepackage{marvosym}
\definecolor{aliceblue}{rgb}{0.94, 0.97, 1.0}
\renewcommand{\thefootnote}{}
\def\paperID{3413} 
\def\confName{CVPR}
\def\confYear{2025}

\title{iSegMan: Interactive Segment-and-Manipulate 3D Gaussians}
\author{
Yian Zhao\textsuperscript{1,3} \ Wanshi Xu\textsuperscript{1} \ Ruochong Zheng\textsuperscript{1,3} \ Pengchong Qiao\textsuperscript{1,3} \ Chang Liu\textsuperscript{4} \ Jie Chen\textsuperscript{1,2,3\ \Letter} \and
\small\textsuperscript{1}School of Electronic and Computer Engineering, Peking University, Shenzhen, China
 \quad
\small\textsuperscript{2}Pengcheng Laboratory, Shenzhen, China \\
\small\textsuperscript{3}AI for Science (AI4S)-Preferred Program, Peking University Shenzhen Graduate School, China \\
\small\textsuperscript{4}Department of Automation and BNRist, Tsinghua University, Beijing, China \\
\small \href{mailto:zhaoyian@stu.pku.edu.cn}{zhaoyian@stu.pku.edu.cn}  \quad \small \href{mailto:jiechen2019@pku.edu.cn}{jiechen2019@pku.edu.cn}
\vspace*{-4mm}
}

\begin{document}
\maketitle
\begin{abstract}
The efficient rendering and explicit nature of 3DGS promote the advancement of 3D scene manipulation.
However, existing methods typically encounter challenges in controlling the manipulation region and are unable to furnish the user with interactive feedback, which inevitably leads to unexpected results.
Intuitively, incorporating interactive 3D segmentation tools can compensate for this deficiency. Nevertheless, existing segmentation frameworks impose a pre-processing step of scene-specific parameter training, which limits the efficiency and flexibility of scene manipulation.
To deliver a 3D region control module that is well-suited for scene manipulation with reliable efficiency, we propose \textbf{i}nteractive \textbf{Seg}ment-and-\textbf{Man}ipulate 3D Gaussians (\textbf{iSegMan}), an interactive segmentation and manipulation framework that only requires simple 2D user interactions in any view.
To propagate user interactions to other views, we propose Epipolar-guided Interaction Propagation (\textbf{EIP}), which innovatively exploits epipolar constraint for efficient and robust interaction matching.
To avoid scene-specific training to maintain efficiency, we further propose the novel Visibility-based Gaussian Voting (\textbf{VGV}), which obtains 2D segmentations from SAM and models the region extraction as a voting game between 2D Pixels and 3D Gaussians based on Gaussian visibility.
Taking advantage of the efficient and precise region control of EIP and VGV, we put forth a \textbf{Manipulation Toolbox} to implement various functions on selected regions, enhancing the controllability, flexibility and practicality of scene manipulation.
Extensive results on 3D scene manipulation and segmentation tasks fully demonstrate the significant advantages of iSegMan.
Project page is available at \href{https://zhao-yian.github.io/iSegMan}{https://zhao-yian.github.io/iSegMan}.
\end{abstract}
\footnote{\Letter\ Corresponding author.}
\vspace*{-8mm}
\section{Introduction}
\label{sec:intro}

\begin{figure}[t]
\hsize=\linewidth
\centering
\includegraphics[width=\linewidth]{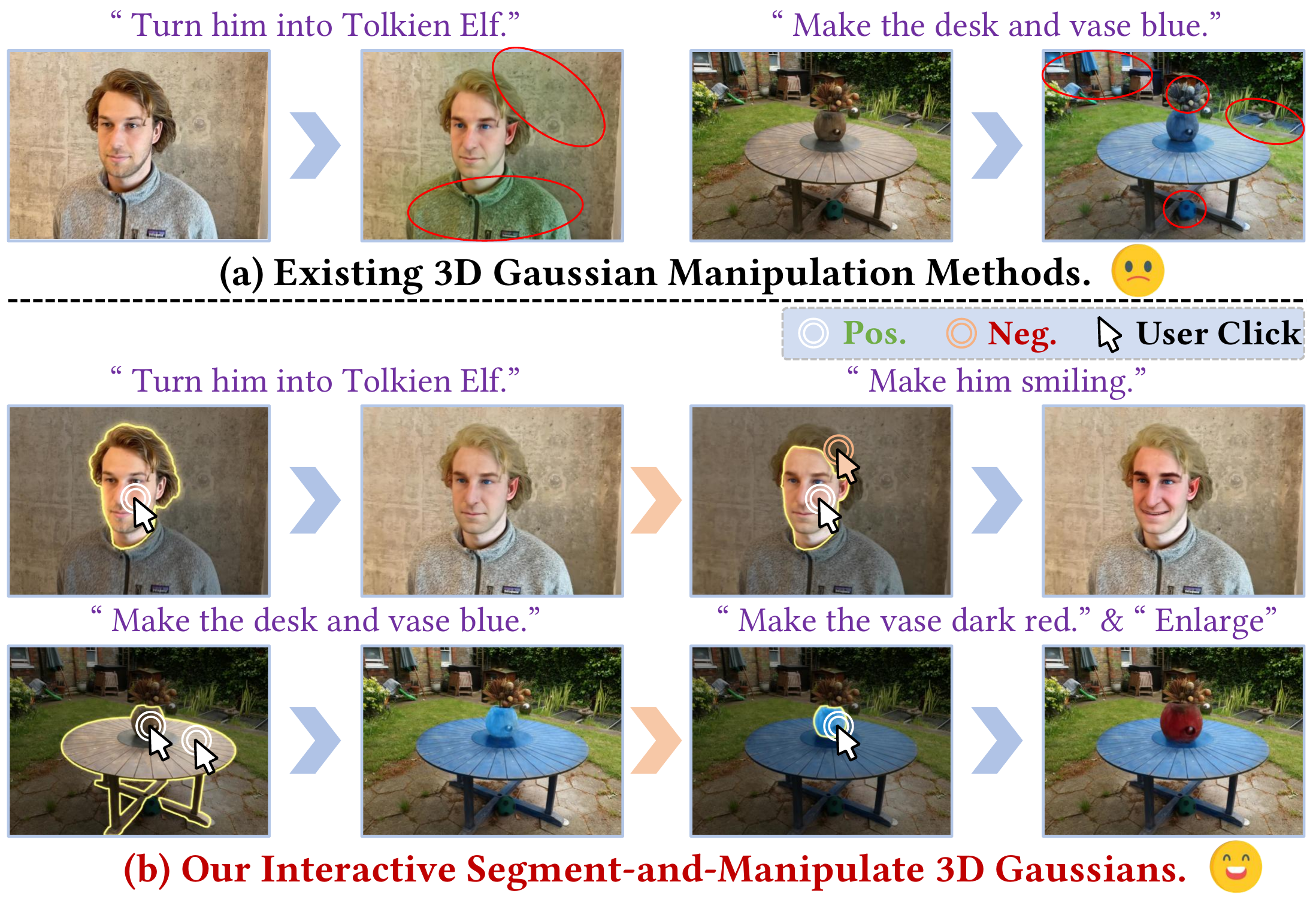}
\vspace{-6mm}
\caption{
\textbf{(a)}: Existing 3D manipulation methods. The red circles mark the irrelevant regions affected by editing, leading to unexpected results.
\textbf{(b)}: Our iSegMan achieves precise control of the manipulation region and interactively performs various functions.
}
\vspace{-6mm}
\label{fig:motivation}
\end{figure}

The capacity to interact with the 3D environments is a critical component across a range of applications, including augmented reality~(AR)~\cite{carmigniani2011augmented}, embodied AI~\cite{duan2022survey}, and spatial computing~\cite{shekhar2015spatial}. 
The advancement of these applications continues to propel innovation in user experience.
Recently, the efficient differentiable rendering and explicit nature of 3D Gaussian Splatting~(3DGS)~\cite{kerbl20233d} have propelled the field of 3D scene manipulation to new frontiers.
However, existing methods typically face challenges in precisely controlling the manipulation region and are unable to provide interactive feedback to users, which inevitably leads to unexpected or uncontrolled results in practice, \textit{cf.}~\cref{fig:motivation}(a).

Intuitively, the above deficiency can be compensated for by incorporating interactive 3D segmentation tools, which accept various types of user interactions to achieve precise control of the manipulation region.
Traditional 3D representations~(\eg,point clouds~\cite{guo2020deep} and meshes~\cite{yao20183d}) typically require users to interact directly in 3D space, which involves complex transformation or post-processing, resulting in a poor user experience.
With the advent of differentiable rendering techniques (\ie, NeRF~\cite{mildenhall2021nerf} and 3DGS~\cite{kerbl20233d}), several interactive 3D segmentation frameworks~\cite{cen2023segment,hu2024semantic,cen2023saga} based on 2D user interaction have been explored, which exploits a priori knowledge of the promptable image segmentation model SAM~\cite{kirillov2023segment} to achieve 3D region selection.
However, these methods usually impose a pre-processing step of scene-specific parameter training, which limits the efficiency and flexibility of 3D scene manipulation.

To deliver a 3D region control module that is well-suited for 3D scene manipulation with reliable efficiency, we propose \textbf{i}nteractive \textbf{Seg}ment-and-\textbf{Man}ipulate 3D Gaussians~(\textbf{iSegMan}), which supports efficient and precise region control and powerful 3D manipulation capability in an interactive manner.
To facilitate user interaction, we first classify the existing 3D interactions into three categories: \textit{3D Click}, \textit{2D Scribble}, and \textit{2D Click}, and elaborate on their characteristics~(see details in~\cref{subsec:3d interaction}).
Considering the simplicity and flexibility of the 2D Click, we adopt it for our framework and permit users to interact from any viewpoint.
To propagate user interactions to other views, we propose Epipolar-guided Interaction Propagation (EIP), which innovatively exploits epipolar constraint for efficient and robust interaction matching.
To avoid scene-specific training to maintain efficiency, we further propose novel Visibility-based Gaussian Voting (VGV), which obtains 2D segmentations from SAM~\cite{kirillov2023segment} and then models the region extraction process as a voting game between \textit{2D Pixels} and \textit{3D Gaussians} based on Gaussian visibility.
Taking advantage of the efficient and precise region control of EIP and VGV, we develop a manipulation toolbox to implement various functions on selected regions, including \textit{Semantic Editing}, \textit{Colorize}, \textit{Scaling}, \textit{Copy\&Paste}, \textit{Combination}, and \textit{Removal}, which significantly enhances the controllability, flexibility and practicality of 3D scene manipulation, \textit{cf.}~\cref{fig:motivation}(b).

To validate the effectiveness of the proposed iSegMan, we perform comprehensive qualitative and quantitative experiments on 3D scene manipulation and segmentation tasks across different scenes, covering all functions provided by the manipulation toolbox.
Our iSegMan not only enables efficient and precise control of the manipulation region, but also supports the progressive editing of complex requirements in an interactive manner and improved reusability of 3D assets.
Moreover, iSegMan achieves the optimal balance of performance and execution speed and excellent robustness in interactive 3D segmentation.

The main contributions can be summarized as:
(\romannumeral1). We propose iSegMan, which precisely controls the manipulation region based on user interactions and invokes functions from the equipped manipulation toolbox according to user requirements, overcoming the limitations of existing methods in controlling the manipulation region and failing to provide interactive feedback to the user.
(\romannumeral2). Two novel algorithms, namely EIP and VGV, are proposed to achieve 3D region segmentation without introducing any scene-specific training, achieving optimal execution speed and accuracy, making them well-suited for scene manipulation.
(\romannumeral3). The proposed manipulation toolbox encompasses versatile inspiring functions, providing a powerful solution for various 3DGS-based applications.
(\romannumeral4). The proposed iSegMan not only provides an efficient and novel solution for interactive 3D segmentation, but also greatly enhances the controllability, flexibility and practicality of 3D scene manipulation.
\section{Related Work}
\label{sec:related}

\subsection{3D Scene Manipulation}
3D scene manipulation is a highly practical application that has received considerable attention from the community.
Recently, 3D manipulation has been implemented mainly based on NeRF~\cite{mildenhall2021nerf} and 3DGS~\cite{kerbl20233d} as follows:

\noindent \textbf{NeRF-based.}
EditNeRF~\cite{liu2021editing} enables the manipulation of the shape and color of the neural fields by conditioning them on latent codes.
CLIP-NeRF~\cite{wang2022clip} and TextDeformer~\cite{gao2023textdeformer} employ the CLIP~\cite{radford2021learning} model to facilitate manipulation through the use of text prompts or reference images.
NeRF-Editing\cite{yuan2022nerf} and NeuMesh~\cite{yang2022neumesh} enable the manipulation of NeRF by converting implicit NeRF representations into explicit meshes and exploiting controllable mesh deformations.
Instruct-N2N~\cite{haque2023instruct}, DreamEditor~\cite{zhuang2023dreameditor}, and GenN2N~\cite{liu2024genn2n} leverage the power of 2D image editors to perform semantic editing on NeRF and achieve impressive results.
However, these NeRF-based methods are limited by the intrinsic complexity of the implicit scene data encoding, making it difficult to control the manipulation region.

\noindent \textbf{3DGS-based.} The inherently explicit nature of 3DGS makes it easy to implement scene manipulation for specific regions.
GSEdit~\cite{palandra2024gsedit} implements global editing of 3D objects, and lacks control over the local region.
\cite{wang2024gaussianeditor} works with LLMs~\cite{liu2023visual,zhuang2025vasparse,zhuang2025vargpt} to provide an automated pipeline and uses existing interactive 3D segmentation tools for additional scene-specific training to control the editing region.
GaussianEditor~\cite{chen2024gaussianeditor} achieves text-driven semantic editing by densifying and optimizing 3D Gaussians within dynamic semantic regions.
Although it supports region control based on text prompts, it is limited by the complexity of text descriptions and lacks interactive capability, making it difficult to segment fine-grained regions.
In contrast, our method provides efficient and precise region control for scene manipulation in an interactive manner.

\begin{figure*}[ht]
\hsize=\linewidth
\centering
\includegraphics[width=0.98\linewidth]{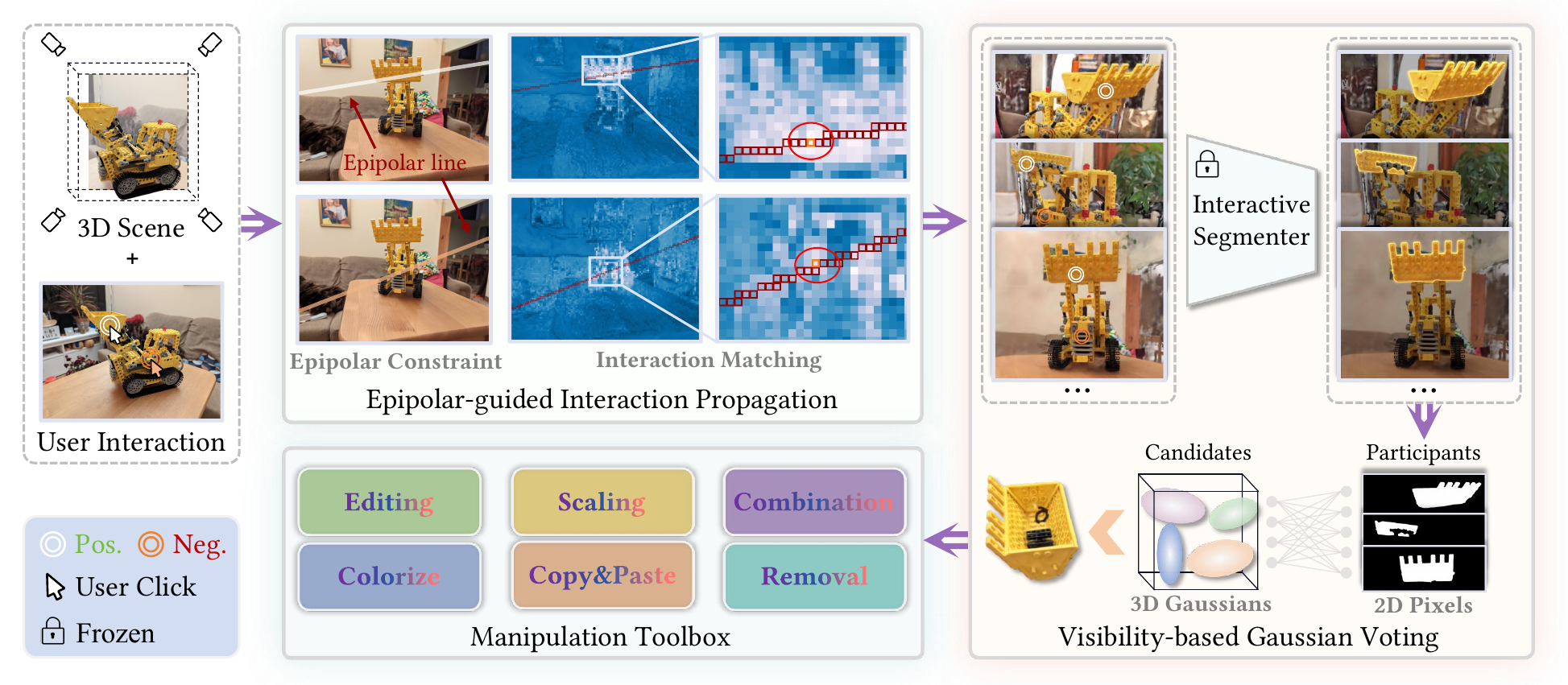}
\vspace{-4mm}
\caption{
\textbf{Overview of iSegMan.}
iSegMan contains two novel region control algorithms that are well-suited for scene manipulation with reliable efficiency: Epipolar-guided Interaction Propagation~(EIP) and Visibility-based Voting Game~(VGV), and a Manipulation Toolbox that includes various manipulation functions.
EIP accepts 2D user interactions in any view and leverages epipolar constraint to efficiently and robustly propagate user interactions to other views.
To avoid scene-specific training to maintain efficiency, VGV obtains 2D mask from SAM and then models the 3D region extraction as a voting game between 2D Pixels and 3D Gaussians based on Gaussian visibility.
Based on the versatile manipulation functions, iSegMan greatly enhances the controllability, flexibility and practicality of 3D scene manipulation.
}
\vspace{-4mm}
\label{fig:overview}
\end{figure*}

\subsection{Interactive 3D Segmentation}
\label{subsec:3d interaction}
Interactive 3D segmentation has been widely used in downstream tasks due to its flexibility and practicality.
Existing methods usually adopt different types of interactions. 
To facilitate analysis the strengths and weaknesses of various interactions, we classify the existing methods according to the interaction type as follows:

\noindent \textbf{3D Click.}
InterObject3D~\cite{kontogianni2023interactive} first develops the interactive 3D segmentation based on point clouds, allowing users to iteratively input positive / negative 3D clicks to interact with the point clouds.
AGILE3D~\cite{yue2023agile3d} efficiently achieves segmentation of multiple objects in the point clouds and also supports multi-round interactions driven by positive / negative 3D clicks.
UniSeg3D~\cite{xu2024unified} unifies multiple 3D segmentation tasks, where interactive segmentation is achieved by 3D superpoints, but this approach only supports positive clicks.
iSeg~\cite{lang2024iseg} proposes the Mesh Feature Field to implement mesh-based interactive segmentation and receive 3D positive / negative clicks on the surface of objects.

\noindent \textbf{2D Scribble.}
NVOS~\cite{ren2022neural} introduces custom-designed 3D features and trains a MLP to achieve scribble-style 3D interaction. 
ISRF~\cite{goel2023interactive} introduces additional feature fields and employs the self-supervised pretrained model to distill semantic features. It extracts 3D regions matching 2D scribble based on feature similarity.
Both require time- and memory-consuming scene-specific feature training.

\noindent \textbf{2D Click.}
Existing methods of this type are typically based on the SAM~\cite{kirillov2023segment}, which provides great potential for interactive 3D segmentation.
SA3D~\cite{cen2023segment} segments 3D objects according to 2D clicks in the initial view by alternating mask inverse rendering and heuristic cross-view self-prompting.
\cite{huang2023point} adopts the same cross-view self-prompting strategy and introduces a two-stage mask refinement scheme. Both methods require multiple repetitions of inverse rendering and involve back-propagation to train the predefined 3D mask in each interaction.
Another line of research is essentially 3D clustering, including OmniSeg3D~\cite{ying2024omniseg3d}, Gaussian Grouping~\cite{ye2023gaussian}, SAGA~\cite{cen2023saga}, LangSplat~\cite{qin2024langsplat}, GARField~\cite{kim2024garfield}, and Click-Gaussian~\cite{choi2024click}. They first utilize SAM to obtain a set of masks for all views~(a time-consuming process), and then distill 3D semantic features from these 2D masks.
Once trained, the semantic feature can be clustered to extract the target 3D object. These methods lack the ability to perform multi-round positive and negative interactions, typically only allow clustering of similar features based on positive clicks, and require time- and memory-consuming data pre-processing and feature training pipelines.

Of these interaction types, 2D Click provides the most concise user interface, and avoids the complex transformation involved with 3D Click.
Consequently, our method adopts 2D Click for interaction and allows users to input in any view. 
Compared with existing methods, our method avoids any scene-specific training, achieving optimal execution speed and accuracy.

\section{Method}

In this section, we elaborate on the proposed iSegMan, which comprises two  pivotal algorithms for region control that are well-suited for scene manipulation with reliable efficiency: Epipolar-guided Interaction Propagation~(EIP) and Visibility-based Voting Game~(VGV), as well as a powerful manipulation toolbox that enables the execution of diverse suite of functions on selected regions~\textit{cf.}~\cref{fig:overview}. 
Specifically, EIP accepts 2D user interactions in any view and leverages epipolar constraint to efficiently and robustly propagate user interactions to other views.
To avoid scene-specific training to maintain efficiency, VGV obtains 2D mask from SAM and then models the 3D region extraction process as a voting game between 2D Pixels and 3D Gaussians based on Gaussian visibility.
Based on the versatile functions of the manipulation toolbox, iSegMan greatly enhances the controllability, flexibility and practicality of 3D scene manipulation.
The details are described below.

\subsection{Epipolar-guided Interaction Propagation}

The EIP is predicated on the principles of Multi-View Stereo~(MVS)~\cite{seitz2006comparison} technology and consists of two steps: epipolar constraint and interaction matching.
Formally, let $\boldsymbol{p}_v = (x_v, y_v)$ represent the coordinates of a user-provided 2D click at the viewpoint $v$.
To propagate $\boldsymbol{p}_v$ to other views, an intuitive idea is to match the image features of other views to the feature at $\boldsymbol{p}_v$.
However, the large search space renders the matching process vulnerable to noise, leading to inefficiency and a lack of robustness in the results.
To address this issue, we introduce the epipolar constraint to restrict the search space.

\begin{figure}[t]
\hsize=\linewidth
\centering
\includegraphics[width=\linewidth]{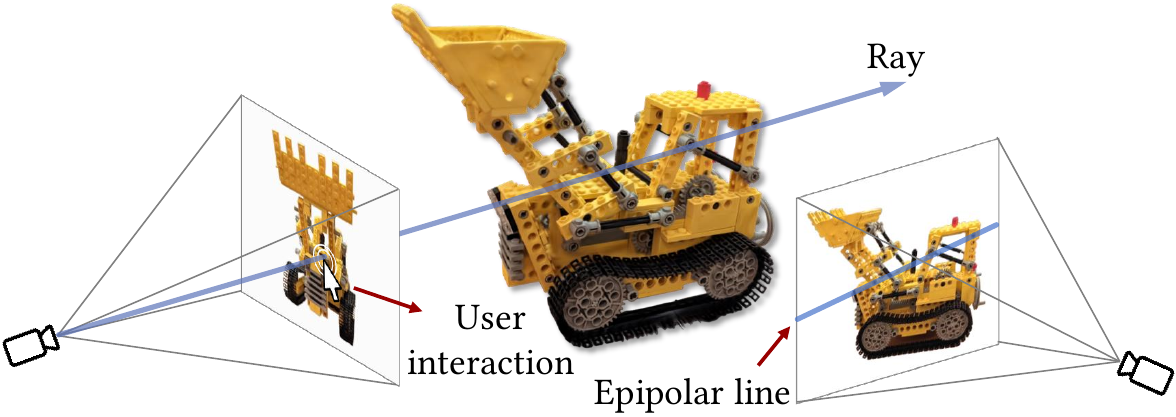}
\vspace{-4mm}
\caption{
\textbf{Illustration of the epipolar constraint.}
}
\vspace{-4mm}
\label{fig:epipolar}
\end{figure}

\noindent \textbf{Epipolar Constraint.}
Since the depth $d_{\boldsymbol{p}_v}$ is a variable when the 2D click $\boldsymbol{p}_v$ is projected into 3D space, this results in a ray $\boldsymbol{r}_{\boldsymbol{p}_v}$ in 3D space that originates from the camera center at the viewpoint $v$.

\noindent \textbf{Theorem 1.} \textit{
$\boldsymbol{r}_{\boldsymbol{p}_v}$ is projected onto an epipolar line $\boldsymbol{e}_{\boldsymbol{p}_v}^{\tilde{v}}$ at each new viewpoint $\tilde{v}$, and the matching click $\boldsymbol{p}_{\tilde{v}}$ must lie on the epipolar line $\boldsymbol{e}_{\boldsymbol{p}_v}^{\tilde{v}}$.
}

\noindent \textit{Proof.} 
Drawing from principles of epipolar geometry~\cite{hartley2003multiple}, the virtual 3D click, whether on the surface or within the 3D object, must lie on the ray $\boldsymbol{r}_{\boldsymbol{p}_v}$.
Consequently, the matching 2D click $\boldsymbol{p}_{\tilde{v}}$ at the new viewpoint $\tilde{v}$ must lie on the epipolar line $\boldsymbol{e}_{\boldsymbol{p}_v}^{\tilde{v}}$, as depicted in~\cref{fig:epipolar}.

Next, we detail the calculation process of the epipolar line $\boldsymbol{e}_{\boldsymbol{p}_v}^{\tilde{v}}$.
Given the camera pose $\boldsymbol{\pi}_v=\mathbf{K}_v[\mathbf{R}_v | \mathbf{t}_v]$, where $\mathbf{K}_v$ and $[\mathbf{R}_v | \mathbf{t}_v]$ are the intrinsic and extrinsic of the camera respectively.
To register the ray $\boldsymbol{r}_{\boldsymbol{p}_v}$ in the world coordinate system, we select two virtual 3D points $\boldsymbol{p}_v^{w_1}$ and $\boldsymbol{p}_v^{w_2}$ on $\boldsymbol{r}_{\boldsymbol{p}_v}$ by sampling the depth $d_{\boldsymbol{p}_v}$, as calculated in~\cref{equ:pw}.
\begin{equation}
\label{equ:pw}
\begin{aligned}
\relax [\mathbf{R}_v | \mathbf{t}_v] = &
    \begin{pmatrix}
        \mathbf{R}_v & \mathbf{t}_v \\
        \overrightarrow{\mathbf{0}}^\mathrm{T} & 1 \\
    \end{pmatrix}, \\
\boldsymbol{p}_v^w = \mathbf{R}_v^{-1}(d_{\boldsymbol{p}_v}&\mathbf{K}_v^{-1} \cdot [\boldsymbol{p}_v^\mathrm{T}, 1]^\mathrm{T} - \mathbf{t}_v).
\end{aligned}
\end{equation}
For simplicity, we set $d_{\boldsymbol{p}_v}$ to 0 and 1 respectively, so $\boldsymbol{p}_v^{w_1}$ and $\boldsymbol{p}_v^{w_2}$ are expressed as~\cref{equ:pw12}.
\begin{equation}
\label{equ:pw12}
\begin{aligned}
    \boldsymbol{p}_v^{w_1}&=-\mathbf{R}_v^{-1}\mathbf{t}_v, \\
    \boldsymbol{p}_v^{w_2}=\mathbf{R}_v^{-1}&(\mathbf{K}_v^{-1} \cdot [\boldsymbol{p}_v^\mathrm{T}, 1]^\mathrm{T} - \mathbf{t}_v).
\end{aligned}
\end{equation}
Finally, we calculate the normalized direction vector $\boldsymbol{\tau}_{\boldsymbol{p}_v}$ of the ray $\boldsymbol{r}_{\boldsymbol{p}_v}$ according to~\cref{equ:dir}.
\begin{equation}
\label{equ:dir}
    \boldsymbol{\tau}_{\boldsymbol{p}_v}=\frac{\boldsymbol{p}_v^{w_1}-\boldsymbol{p}_v^{w_2}}{\Vert\boldsymbol{p}_v^{w_1}-\boldsymbol{p}_v^{w_2}\Vert}.
\end{equation}

To calculate the epipolar line $\boldsymbol{e}_{\boldsymbol{p}_v}^{\tilde{v}}$ in the camera coordinate system of the new viewpoint $\tilde{v}$, it is sufficient to transform the coordinate system of the registered ray $\boldsymbol{r}_{\boldsymbol{p}_v}$ again using the camera pose $\boldsymbol{\pi}_{\tilde{v}}=\mathbf{R}_{\tilde{v}}[\mathbf{R}_{\tilde{v}} | \mathbf{t}_{\tilde{v}}]$.
Similarly, we sample two virtual 3D points from $\boldsymbol{r}_{\boldsymbol{p}_v}$ for the transformation, and the corresponding 2D points $\boldsymbol{p}_v^{\tilde{v}}$ in the camera coordinate system of the viewpoint $\tilde{v}$ are calculated as~\cref{equ:ep}.
\begin{equation}
\label{equ:ep}
    [{\boldsymbol{p}_v^{\tilde{v}}}^\mathrm{T}, 1]^\mathrm{T} = \frac{1}{d_{\boldsymbol{p}_{\tilde{v}}}}\mathbf{K}_{\tilde{v}}(\mathbf{R}_{\tilde{v}}\boldsymbol{p}_v^w + \mathbf{t}_{\tilde{v}}).
\end{equation}
Utilizing the two points $\boldsymbol{p}_v^{\tilde{v}_1}$ and $\boldsymbol{p}_v^{\tilde{v}_2}$, we are able to precisely derive the expression for the epipolar line $\boldsymbol{e}_{\boldsymbol{p}_v}^{\tilde{v}}$ within the camera coordinate system.

\noindent \textbf{Interaction Matching.}
To acquire the matching 2D click $\boldsymbol{p}_{\tilde{v}}$ at the viewpoint $\tilde{v}$, we further perform the interaction matching based on semantic feature affinity.
Specifically, we utilize the self-supervised pretrained model~(\eg, DINO~\cite{caron2021emerging}) as the feature extractor, where the feature maps of views $\mathcal{I}_v$ and $\mathcal{I}_{\tilde{v}}$ are denoted as $\mathbf{F}_v$ and $\mathbf{F}_{\tilde{v}}$, respectively.
Due to the epipolar constraint, the search space is significantly reduced and we only need to calculate the affinity $\mathcal{A}_{\boldsymbol{p}_v}^{\tilde{v}}$ between the feature $\mathbf{F}_v[\boldsymbol{p}_v] \in \mathbb{R}^{1 \times D}$ and the discontinuous feature sequence $\mathbf{F}_{\tilde{v}}[\boldsymbol{e}_{\boldsymbol{p}_v}^{\tilde{v}}] \in \mathbb{R}^{M \times D}$~($M$ indicates the length of the feature sequence, and $D$ denotes the feature dimension), thus reducing noise errors and improving the accuracy and robustness.
For implementation, inspired by the Bresenham algorithm~\cite{bresenham1998algorithm}, we efficiently gather the discontinuous feature sequence $\mathbf{F}_{\tilde{v}}[\boldsymbol{e}_{\boldsymbol{p}_v}^{\tilde{v}}]$ and corresponding indices $\boldsymbol{I}_{\tilde{v}}$ along the epipolar line $\boldsymbol{e}_{\boldsymbol{p}_v}^{\tilde{v}}$.
Finally, we upsample the coordinates of the selected feature vector with the highest affinity to the original view size to obtain the coordinates of matching 2D click $\boldsymbol{p}_{\tilde{v}}$, \textit{cf.}~\cref{equ:matching}.
\begin{equation}
\begin{split}
\label{equ:matching}
    & \mathcal{A}_{\boldsymbol{p}_v}^{\tilde{v}} = \mathbf{F}_v[\boldsymbol{p}_v] \cdot {\mathbf{F}_{\tilde{v}}[\boldsymbol{e}_{\boldsymbol{p}_v}^{\tilde{v}}]}^T \in \mathbb{R}^{1 \times M}, \\
    &\boldsymbol{p}_{\tilde{v}}=\texttt{Upsample}(\boldsymbol{I}_{\tilde{v}}[\texttt{argmax}(\mathcal{A}_{\boldsymbol{p}_v}^{\tilde{v}})]).
\end{split}
\end{equation}

\subsection{Visibility-based Gaussian Voting}
Based on the interactions of all the views obtained by EIP, we employ the SAM~\cite{kirillov2023segment} to obtain a set of 2D binarized masks $\boldsymbol{\mathcal{M}}=\{\mathbf{m}_i | \mathbf{m}_i \in \{0, 1\}^{h \times w}\}_{i=1}^{K}$, where $K$ denotes the number of views, $1$ means the pixel is rendered by the target region and $0$ means the pixel is rendered by the irrelevant region, $h$ and $w$ are the height and width of the views, respectively.
Our goal is to extract target 3D Gaussians from the entire scene based on $\boldsymbol{\mathcal{M}}$.
To avoid scene-specific training to maintain efficiency, we model the region extraction process as a voting game from \textit{2D Pixels} to \textit{3D Gaussians}.

\noindent \textbf{Voting Principle.}
Voting involves a two-party game, namely the participants and the candidates.
We treat 2D Pixels as the participant set $\boldsymbol{P}$ and 3D Gaussians as the candidate set $\boldsymbol{C}$.
There are a total of $h \times w$ participants and $N$ candidates, where $N$ is the number of 3D Gaussians contained in the entire scene.
Based on the set of 2D masks $\boldsymbol{\mathcal{M}}$, each participant $\boldsymbol{p}_i \in \boldsymbol{P}$ is assigned a vector $\boldsymbol{\tau}_i = ( t_1, t_2, \dots, t_K )$, where $t_k \in \{0, 1\}$ for all $k$, to indicate whether the visible 3D Gaussians belong to the target region from $K$ views.
\noindent \textbf{Theorem 2.} \textit{The voting of 2D Pixels on 3D Gaussians is cumulative and asymmetric.}

\noindent \textit{Proof.}
(\romannumeral1). \textit{Cumulative}: each participant $\boldsymbol{p}_i$ is allowed to vote $K$~($K > 1$) times, \ie, once for each view, so voting is cumulative.
(\romannumeral2). \textit{Asymmetric}: each participant $\boldsymbol{p}_i$ has different voting powers for different candidates, as each 2D Pixel has a different degree of visibility to 3D Gaussians at distinct positions and depths.
Intuitively, the higher the visibility of a candidate to a participant, the higher the probability that the candidate belongs to the same category as the participant (inside or outside the target region). Conversely, the higher the degree of occlusion of a candidate to a participant, the more uncertain the participant is about the candidate and the voting power is reduced.

Inspired by the \textit{Alpha Blending} of colors in splatting rendering~\cite{kerbl20233d}, we define the voting power $\Upsilon_{i,j}$ of each participant $\boldsymbol{p}_i$ for each candidate $\boldsymbol{c}_j$ as the \textit{Alpha Blending} of its visibility (the opacity of 3D Gaussians), as calculated in~\cref{equ:power}.
The detailed technical principle of 3DGS~\cite{kerbl20233d} and the calculation of $\alpha$ are presented in the Appendix 4.
\begin{equation}
\label{equ:power}
    \Upsilon_{i,j} = \sigma_i \cdot \alpha_i \prod_{k=1}^{i-1} (1-\alpha_k).
\end{equation}
Once the voting power has been determined, all participants can vote for all candidates and the number of votes for each candidate is calculated according to~\cref{equ:vote}.
\begin{equation}
\label{equ:vote}
    \Psi_j = \frac{1}{h \times w \times K} \cdot \sum_{i} \sum_{k} \boldsymbol{\tau}_i[k] \cdot \Upsilon_{i,j}.
\end{equation}
Finally, we select the candidates (3D Gaussians) with the number of votes greater than the predetermined threshold to accurately extract the target region.

\noindent \textbf{Iterative Inspection Mechanism.}
In the context of open-world scenes, the target region may be invisible at certain viewpoints due to occlusion or out-of-view, resulting in erroneous 2D segmentations produced by SAM.
To address this issue, we propose the Iterative Inspection Mechanism~(IIM).
Specifically, we iteratively execute the voting process at each viewpoint $v$ to obtain the currently selected 3D Gaussians and render the corresponding 2D rendered mask $\mathbf{m}_v^r$ of that view.
If the mask $\mathbf{m}_v^p$ predicted by SAM in this view does not intersect with the rendered mask $\mathbf{m}_v^r$, IIM determines that the target region cannot be observed at viewpoint $v$ and does not retain the predicted mask $\mathbf{m}_v^p$.
Furthermore, the IIM is capable of mitigating the potential for noise errors introduced by EIP and the SAM.
As each predicted mask $\mathbf{m}_v^p$ must be reviewed by the IIM prior to being allowed to participate in the voting process, any incorrect matching interactions or anomalous segmenter behaviour will be excluded, thus enhancing the robustness.
It is worth noting that the implementation of millisecond-level 3D Gaussian voting and rendering ensures that the impact of the IIM on execution speed is negligible.

\begin{figure*}[ht]
\hsize=\linewidth
\centering
\includegraphics[width=\linewidth]{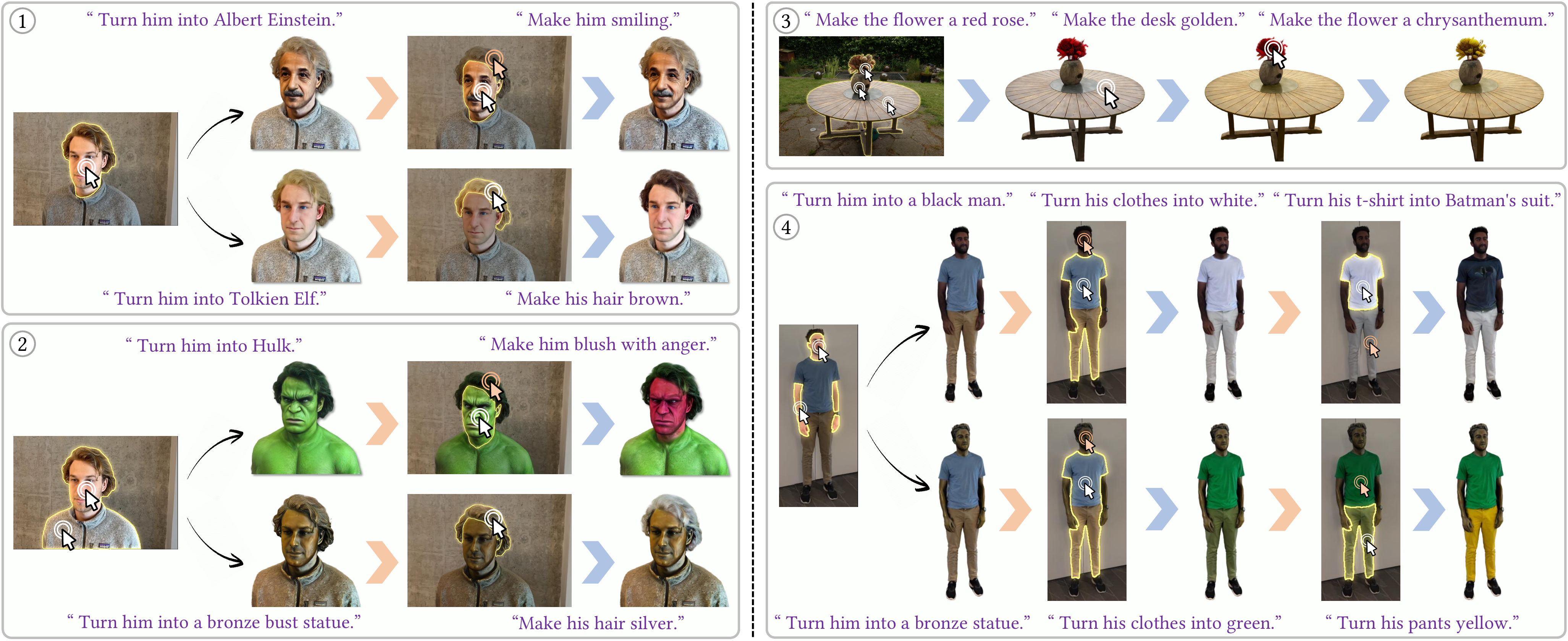}
\vspace{-5mm}
\caption{
\textbf{Results of semantic editing.}
Orange arrows indicate interactive 3D segmentation, and blue arrows indicate semantic editing.
}
\vspace{-5mm}
\label{fig:editing}
\end{figure*}
\begin{figure}[ht]
\hsize=\linewidth
\centering
\includegraphics[width=\linewidth]{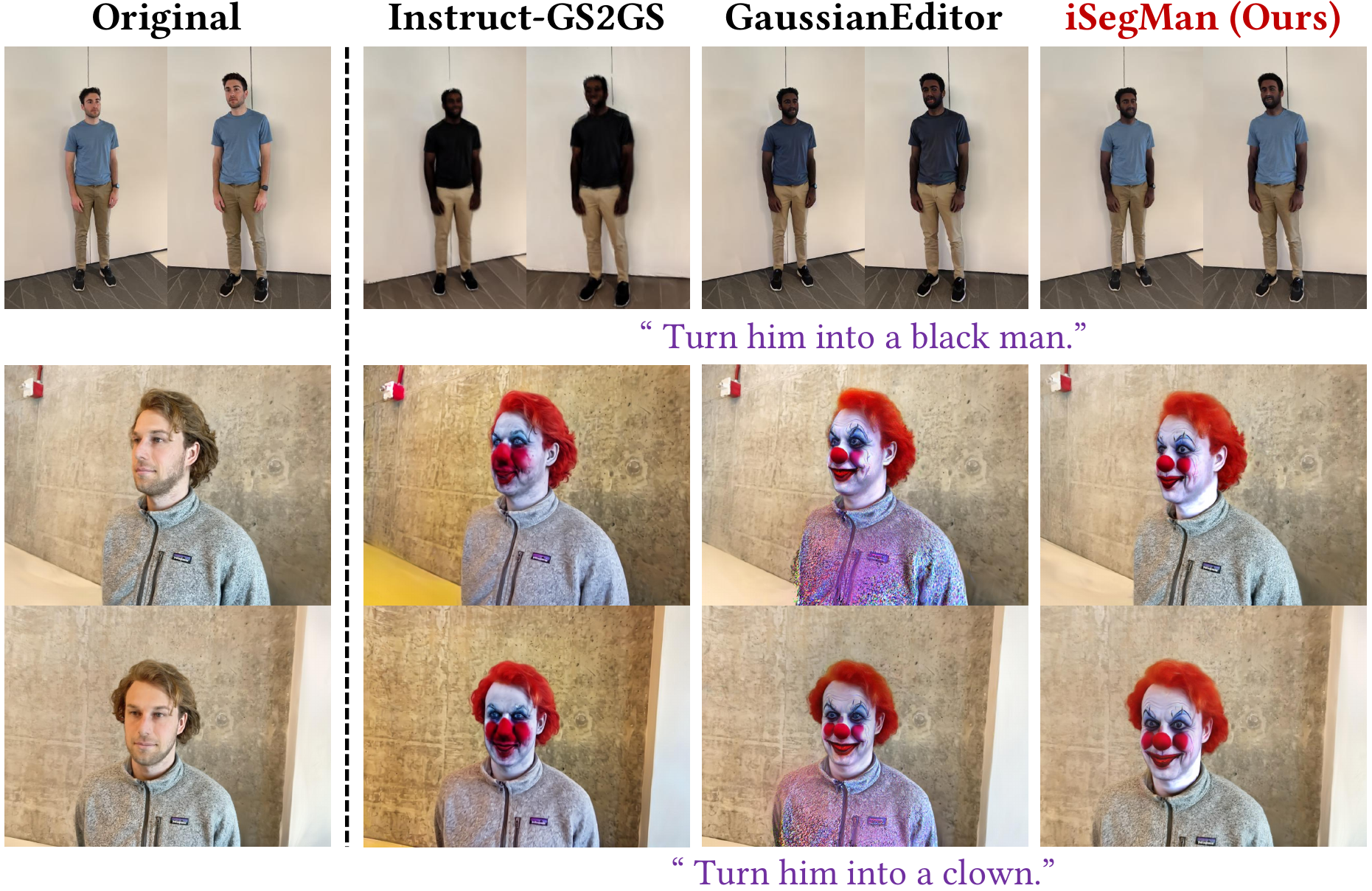}
\vspace{-4mm}
\caption{
\textbf{Comparison of semantic editing.}
}
\vspace{-4mm}
\label{fig:comparison}
\end{figure}

\subsection{Manipulation Toolbox}
Taking advantage of the efficient and precise region control of EIP and VGV, we put forth a Manipulation Toolbox to implement various functions on selected regions. These functions are detailed below.

\noindent \textbf{Semantic Editing.}
This function refers to text-driven editing according to the instruction provided by the user.
We leverage a powerful image editor, InstructPix2Pix~\cite{brooks2023instructpix2pix}, to edit the rendered views and iteratively update the 3D Gaussians using the difference between the edited and original views to achieve 3D editing, following~\cite{haque2023instruct,chen2024gaussianeditor}.
Specifically, we denote the original scene represented by 3D Gaussians as $\Theta$, and the selected region as $\Theta_s$.
$\Theta_s$ is a non-empty subset of $\Theta$, \ie, $\Theta_s \subseteq \Theta \wedge \Theta_s \neq \emptyset $. 
Given a set of viewpoints $\boldsymbol{V}$ of a scene, we first use the differentiable renderer $\mathcal{R}$ to get the rendered image $\mathcal{I}_v$ at each viewpoint $v \in \boldsymbol{V}$.
Then, we iteratively update the 3D Gaussians to maintain the multi-view consistency.
In each iteration, we randomly sample a view $\mathcal{I}_v$ and employ the image editor $\mathcal{E}$ to edit $\mathcal{I}_v$ based on the instruction $e$ to obtain $\mathcal{I}_v^e$.
Finally, the image-level loss between $\mathcal{I}_v$ and $\mathcal{I}_v^e$ is calculated to update $\Theta_s$.
The calculation process is shown in~\cref{equ:editing} and \cref{equ:update}.
\begin{equation}
\label{equ:editing}
\mathcal{I}_v = \mathcal{R}(\Theta, v), \ \ 
\mathcal{I}_v^e = \mathcal{E}(\mathcal{I}_v, e),
\end{equation}
\begin{small}
\begin{equation}
\label{equ:update}
\nabla_\theta \Theta_s = \mathbb{E}_v\left[\left( \frac{\partial \Vert \mathcal{I}_v^e - \mathcal{I}_v \Vert_1}{\partial \mathcal{I}_v} + \frac{\partial \mathcal{D}(\mathcal{I}_v, \mathcal{I}_v^e)}{\partial \mathcal{I}_v} \right) \cdot \frac{\partial \mathcal{I}_v}{\partial \theta} \right],
\end{equation}
\end{small}

\noindent where $\theta$ denotes the trainable parameters of the 3D Gaussians contained in $\Theta_s$, $\mathcal{D}(\cdot, \cdot)$ represents the perceptual distance~\cite{zhang2018unreasonable}.
Note that semantic editing requires multi-step parameter updates, resulting in additional time consumption, but this is not caused by region control.
In addition, an annealing strategy is incorporated into the updating of the 3D Gaussians, where the offset of each step is progressively reduced until it reaches zero. We observe that this operation is beneficial in the editing stability.

\noindent \textbf{Colorization.}
This function changes the color of the selected region by modifying the color attribute of the selected 3D Gaussians.
Specifically, we support two modes: \textit{Color Replacement} and \textit{Balanced Coloring}.
The former is achieved by assigning the color of all selected 3D Gaussians to the target color $\boldsymbol{c}_t$. The latter is achieved by adjusting the mean color value to $\boldsymbol{c}_t$, as calculated in~\cref{equ:color}.
\begin{equation}
\label{equ:color}
\boldsymbol{c}_i = \boldsymbol{c}_i + (\boldsymbol{c}_t - \frac{1}{\hat{N}}\sum_{i=1}^{\hat{N}}\boldsymbol{c}_i),
\end{equation}
where $\hat{N}$ is the number of selected 3D Gaussians.

\begin{figure*}[ht]
\hsize=\linewidth
\centering
\includegraphics[width=\linewidth]{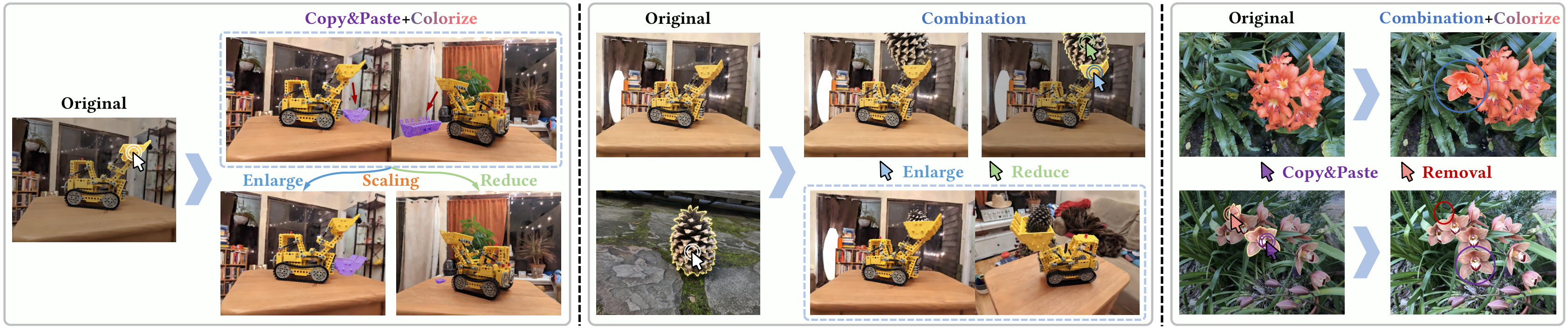}
\vspace{-6mm}
\caption{
\textbf{Results of other manipulation functions.}
}
\vspace{-5mm}
\label{fig:manipulation}
\end{figure*}
\begin{figure}[t]
\hsize=\linewidth
\centering
\includegraphics[width=\linewidth]{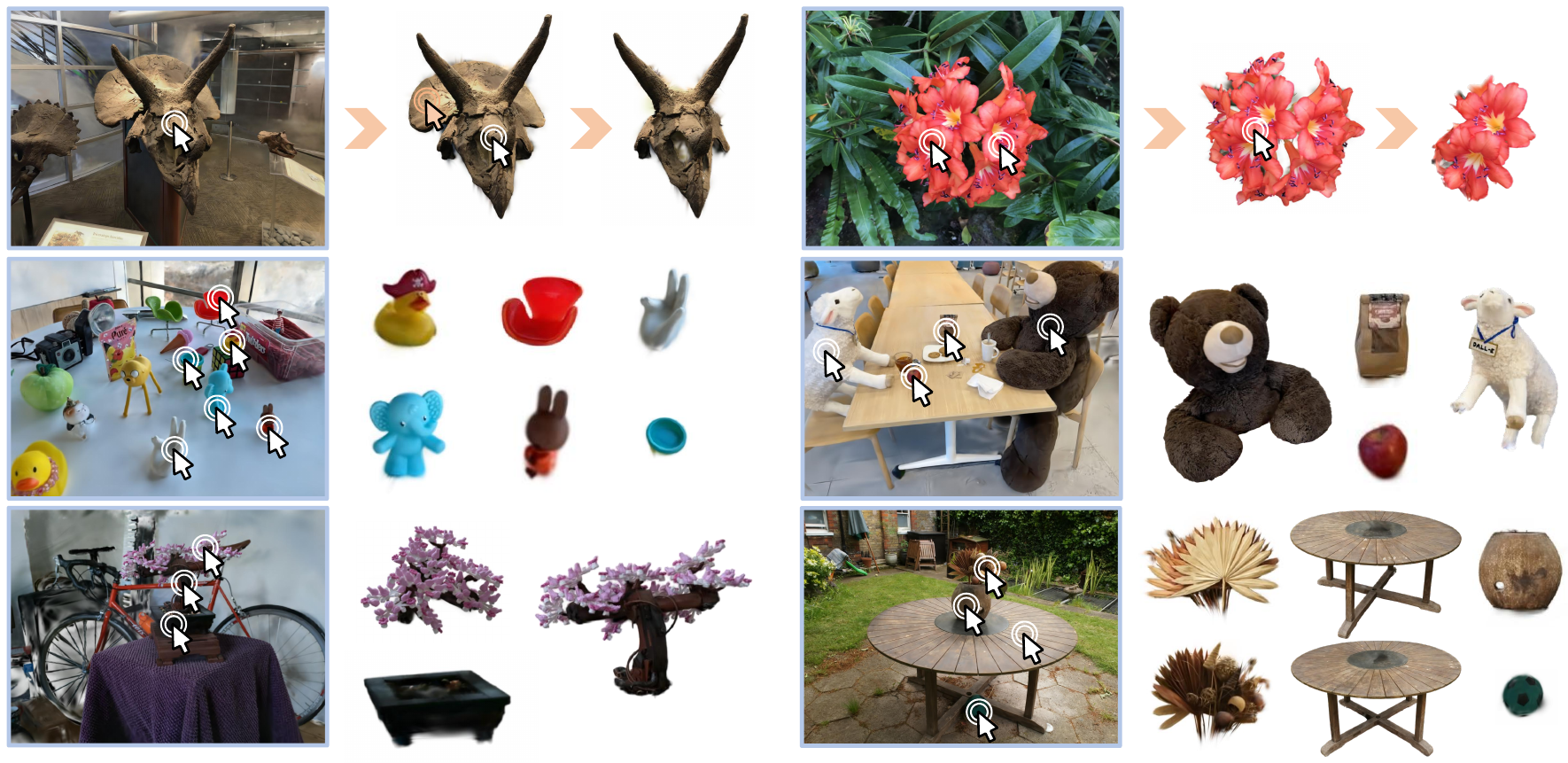}
\vspace{-6mm}
\caption{
\textbf{Visualization of interactive 3D segmentation.}
}
\vspace{-5mm}
\label{fig:segmentation}
\end{figure}

\noindent \textbf{Scaling.}
This function enlarges or reduces the selected region while leaving the rest of the scene unchanged. This is achieved by modifying the scaling factor of the selected 3D Gaussians.
For implementation, the user is allowed to specify a coefficient $\epsilon$ with a value greater than zero.
We first calculate the geometric center of the selected 3D Gaussians and then obtain the direction vector of each 3D Gaussian relative to the geometric center.
To maintain the geometric invariance for rigid transformation, it is imperative that both the direction vector and the scaling factor of each 3D Gaussian be concurrently scaled by the user-specified coefficient.
The calculation is detailed in~\cref{equ:scaling}.
\begin{equation}
\begin{split}
\label{equ:scaling}
\bar{\boldsymbol{\mu}} &= \frac{1}{\hat{N}} \sum_{i=1}^{\hat{N}}\boldsymbol{\mu}_i, \ \ \hat{\boldsymbol{S}}_i = \boldsymbol{S}_i \cdot \epsilon, \\
&\hat{\boldsymbol{\mu}}_i = (\boldsymbol{\mu}_i - \bar{\boldsymbol{\mu}}) \cdot \epsilon + \bar{\boldsymbol{\mu}},
\end{split}
\end{equation}
where $\hat{\boldsymbol{S}}_i$ and $\hat{\boldsymbol{\mu}}_i$ represent the new scaling factor and position of the selected 3D Gaussians, respectively.

\noindent \textbf{Copy\&Paste.}
This function copies the selected region and pastes it elsewhere in the same scene.

\noindent \textbf{Combination.}
This function extracts the selected region in one scene and inserts it into another scene.

\noindent \textbf{Removal.}
This function deletes the selected region.
\section{Experiments}
\subsection{Experimental Settings}

\noindent \textbf{Dataset.}
To demonstrate and compare the performance of 3D manipulation, we perform experiments on two datasets: Mip-NeRF 360~\cite{barron2022mip} and Instruct-N2N~\cite{haque2023instruct}.
For interactive 3D segmentation, we compare quantitative results with existing methods on two commonly used datasets: NVOS~\cite{ren2022neural} and SPIn-NeRF~\cite{mirzaei2023spin}, and further present qualitative results on a sample of scenes on LERF~\cite{kerr2023lerf} and LLFF~\cite{mildenhall2019local}.
See the Appendix 1.1 for a detailed description of the dataset.

\noindent \textbf{Metrics.}
We perform quantitative comparisons of two tasks: semantic editing and interactive 3D segmentation.
For semantic editing, we utilize user study and CLIP direction similarity~\cite{gal2022stylegan} as metrics following~\cite{haque2023instruct,chen2024gaussianeditor}.
For interactive 3D segmentation, we utilize mAcc and mIoU as metrics following previous works~\cite{cen2023segment,cen2023saga}.

\noindent \textbf{Implementation Details.}
All implementation details of the proposed iSegMan are described in the Appendix 1.2.

\subsection{Qualitative Results}

\noindent \textbf{Results of Semantic Editing.}
To demonstrate the advantages of our iSegMan, we first present the semantic editing results on four cases, \textit{cf.}~\cref{fig:editing}.
The user provides 2D clicks and the editing instruction, and iSegMan rapidly extracts the target region based on the 2D clicks and performs editing, which is completed in a few minutes.
This process allows iterative execution in an interactive manner, forming an editing loop until the user requirements are met.
Building such an editing loop presents two distinct advantages.
Firstly, it is an effective way for fulfilling complex editing requirements, \eg, the editing process of Case 4 achieves a complex requirement: ``Turn the person into a bronze statue wearing a green shirt and yellow pants.''
Secondly, it enables reuse of existing results to enhance computational efficiency, \eg, the reuse of the ``golden table'' in Case 3.

\noindent \textbf{Comparison of Semantic Editing.}
Moreover, we qualitatively compare our iSegMan with existing methods Instruct-GS2GS~\cite{haque2023instruct} and GaussianEditor~\cite{chen2024gaussianeditor}, \textit{cf.}~\cref{fig:comparison}.
Since Instruct-GS2GS cannot explicitly control the editing region, irrelevant regions are significantly affected, \eg, the shirt of the person in the first row has become black by mistake, and the wall color in the second row has become darker.
GaussianEditor provides an additional text prompt to specify the editing region. However, the text prompt is difficult to describe for various fine-grained regions, resulting in a poor segmentation accuracy and defective editing results.
For instance, the person's shirts are affected in both scenes, leading to unexpected results.
In contrast, our iSegMan achieves precise region control and excellent editing results.

\noindent \textbf{Results of Other Manipulation Functions.}
We also present the results of other functions in the manipulation toolbox, \textit{cf.}~\cref{fig:manipulation}.
Our iSegMan achieves various functions in an interactive manner, greatly enhancing the controllability, flexibility and practicality of 3D manipulation.

\noindent \textbf{Visualization of Interactive 3D Segmentation.}
To further demonstrate that our iSegMan enables precise region control, we present the visualization of interactive 3D segmentation, \textit{cf.}~\cref{fig:segmentation}.
Our iSegMan accurately segments fine-grained regions based on 2D clicks and requires no scene-specific training, providing a solid foundation for subsequent manipulation tasks.

\subsection{Quantitative Results}

\begin{table}[t]
\centering
\renewcommand{\arraystretch}{1.15}
\setlength{\tabcolsep}{1.91mm}
\footnotesize
\begin{tabular*}{\linewidth}{l | ccc}
\noalign{\hrule height 1.2pt}
Metric & Instruct-GS2GS & GaussianEditor & iSegMan (Ours) \\
\hline
User study $ \uparrow $ & 2.10 $\pm$ 0.20  & 3.32 $\pm$ 0.40 & \textbf{4.52 $\pm$ 0.20} \\
CLIP$_{dir}$ $ \uparrow $ & 0.1647 & 0.2071 & \textbf{0.2189} \\
\noalign{\hrule height 1.2pt}
\end{tabular*}
\vspace{-2mm}
\caption{Quantitative comparison of semantic editing. CLIP$_{dir}$ is the CLIP directional similarity.}
\vspace{-2mm}
\label{tab:editing}
\end{table}

\begin{table}[t]
\centering
\renewcommand{\arraystretch}{1.2}
\setlength{\tabcolsep}{2mm}
\footnotesize
\begin{tabular*}{\linewidth}{l ccccc}
\noalign{\hrule height 1.2pt}
\multirow{2}{*}{Method} & \multirow{2}{*}{Training} & \multirow{2}{*}{\makecell[c]{mIoU\\(\%)}} & \multirow{2}{*}{\makecell[c]{mAcc\\(\%)}} & \multicolumn{2}{c}{Execution Time} \\
\cline{5-6}
& & & & Feature & Segment \\
\hline
MVSeg~\cite{mirzaei2023spin} & \checkmark & 90.4 & {98.8} & - & - \\
ISRF~\cite{goel2023interactive} & \checkmark & 71.5 & 95.5 & - & - \\
SA3D~\cite{cen2023segment} & \checkmark & \underline{91.9} & \underline{98.8} & \underline{5min} & 30s \\
LangSplat~\cite{qin2024langsplat} & \checkmark & 69.5 & 94.5 & \tiny$\sim$\footnotesize2.5h & - \\
SAGA~\cite{cen2023saga} & \checkmark & {88.0} & 98.5 & \tiny$\sim$\footnotesize1.5h & \textbf{10ms} \\
\hline
\rowcolor{aliceblue!80}
iSegMan (Ours) & \textcolor{ForestGreen}{N/A} & \textbf{92.4} & \textbf{99.1} & \textbf{52s} & {6s} \\
\noalign{\hrule height 1.2pt}
\end{tabular*}
\vspace{-2mm}
\caption{Comparison of interactive 3D segmentation on SPIn-NeRF.
``Feature'' column indicates the latency of feature training or extraction, and ``Segment'' column indicates the segmentation latency of each interaction.
}
\vspace{-7mm}
\label{tab:spin}
\end{table}

\noindent \textbf{Comparison of Semantic Editing.}
We perform a user study and calculate the CLIP directional similarity~\cite{gal2022stylegan} to quantitatively compare the performance of semantic editing with existing methods (see the Appendix 2 for evaluation details of both metrics). The results are presented in~\cref{tab:editing}.
iSegMan achieves the optimal performance through flexible and fine-grained control over the editing region.

\noindent \textbf{Comparison of  Interactive 3D Segmentation.}
We compare the performance of interactive 3D segmentation with previous methods on SPIn-NeRF and NVOS datasets, \textit{cf.}~\cref{tab:spin} and \cref{tab:nvos}.
\textbf{Bold} indicates the best performance and \underline{underlined} the second best.
``Feature'' column indicates the latency of feature training or extraction, and ``Segment'' column indicates the segmentation latency of each interaction.
The execution time of some methods is not reported because they do not support segmentation of 3D Gaussians, and the segmentation time at each interaction of LangSplat~\cite{qin2024langsplat} is not reported because it does not support interactive segmentation.
Our iSegMan achieves excellent performance with less execution time and does not require any supervised training with masks.

\begin{table}[t]
\centering
\renewcommand{\arraystretch}{1.2}
\setlength{\tabcolsep}{2mm}
\footnotesize
\begin{tabular*}{\linewidth}{l ccccc}
\noalign{\hrule height 1.2pt}
\multirow{2}{*}{Method} & \multirow{2}{*}{Training} & \multirow{2}{*}{\makecell[c]{mIoU\\(\%)}} & \multirow{2}{*}{\makecell[c]{mAcc\\(\%)}} & \multicolumn{2}{c}{Execution Time} \\
\cline{5-6}
& & & & Feature & Segment \\
\hline
NVOS~\cite{ren2022neural} & \checkmark & 70.1 & 92.0 & - & -\\
ISRF~\cite{goel2023interactive} & \checkmark & 83.8 & 96.4 & - & - \\
SA3D~\cite{cen2023segment} & \checkmark & 90.3 & 98.2 & \underline{2min} & 15s \\
LangSplat~\cite{qin2024langsplat} & \checkmark & 74.0 & 94.0 & \tiny$\sim$\footnotesize2h & - \\
SAGA~\cite{cen2023saga} & \checkmark & \underline{90.9} & \underline{98.3} & \tiny$\sim$\footnotesize1h & \textbf{10ms} \\
\hline
\rowcolor{aliceblue!80}
iSegMan (Ours) & \textcolor{ForestGreen}{N/A} & \textbf{92.0} & \textbf{98.4} & \textbf{30s} & {4s} \\
\noalign{\hrule height 1.2pt}
\end{tabular*}
\vspace{-2mm}
\caption{Comparison of interactive 3D segmentation on NVOS.}
\vspace{-3mm}
\label{tab:nvos}
\end{table}

\subsection{Analysis and Ablation Study}

\begin{table}[t]
\centering
\renewcommand{\arraystretch}{1.2}
\setlength{\tabcolsep}{4.08mm}
\footnotesize
\begin{tabular*}{\linewidth}{c | cccc}
\noalign{\hrule height 1.2pt}
\multirow{2}{*}{\makecell[c]{Sampling \\ Rate}} & \multirow{2}{*}{\makecell[c]{mIoU\\(\%)}} & \multirow{2}{*}{\makecell[c]{mAcc\\(\%)}} & \multicolumn{2}{c}{Execution Time} \\
\cline{4-5}
& & & Feature & Segment \\
\hline
100\% & 92.4 & 99.1 & 52s & 6s \\
100\%$^\clubsuit$ & 92.4 & 99.1 & 52s & 6s \\
50\% & 92.2 & 99.1 & 27s & 4s \\
25\% & 92.1 & 99.0 & 14s & 2s \\
10\% & 92.1 & 99.0 & 7s & 1s \\
\noalign{\hrule height 1.2pt}
\end{tabular*}
\vspace{-2mm}
\caption{Results of robustness analysis. $^\clubsuit$ denotes shuffling the view order.}
\vspace{-5mm}
\label{tab:sparse}
\end{table}

\noindent \textbf{Robustness Analysis.}
To verify the generalization of our iSegMan under different 3D scenes, we perform a robustness analysis.
Specifically, we evaluate the accuracy and execution time of the proposed iSegMan on the SPIn-NeRF dataset under different uniform view sampling rates and shuffled view order~(denoted by $^\clubsuit$) conditions based on the original camera trajectory, \textit{cf.}~\cref{tab:sparse}.
The lower the sampling rate, the worse the coherence between views, and the lower the computational cost, leading to faster execution time.
In addition, shuffling the view order requires segmenting objects from a completely incoherent view list.
The results demonstrate that our iSegMan is capable of maintaining a high level of accuracy, regardless of under sparse and incoherent view conditions (\eg, with a sampling rate of only 10\%), or shuffling of the view order.
Therefore, our iSegMan is highly robust and enables a trade-off between performance and execution time by reducing the view sampling rate.
In contrast, the effectiveness of the cross-view self-prompting strategy proposed by SA3D~\cite{cen2023segment} depends on the accuracy of the rendered mask confidence map, which is limited by the coherence of the rendering viewpoints. 
Moreover, to ensure the stability of the gradient-based training of the 3D mask, SA3D requires that the number of views should not be too few.
Consequently, it is challenging to apply the self-prompting strategy in situations where there is a high degree of visual inconsistency or sparse views.

\noindent \textbf{Ablation Studies.}
We perform ablation studies on the epipolar constraint, the feature extractor, and the iterative inspection mechanism to verify their effectiveness. The results are presented in Tab. B, Tab. C, and Tab. D in Appendix 3 respectively.
The results show that removing the epipolar constraint or the iterative inspection mechanism introduces noise that leads to a significant loss of accuracy, and that our method is robust to the feature extractor.
\section{Conclusion and Limitation}
\paragraph{Conclusion.}
In this paper, we propose a practical interactive AI agent, namely iSegMan, which precisely controls the manipulation region based on user interactions and invokes functions from the equipped manipulation toolbox according to user requirements, overcoming the limitations of existing methods in controlling the manipulation region and providing interactive feedback to the user.
We design two novel algorithms for interactive 3D segmentation that completely avoid the pre-processing step of scene-specific training, making them well-suited for 3D scene manipulation with reliable efficiency and robustness.
The equipped manipulation toolbox encompasses versatile inspiring functions, providing a powerful solution for various 3DGS-based applications.
Extensive experiments show that our iSegMan has significant advantages for interactive 3D segmentation and manipulation tasks.
We hope that our iSegMan will serve as a practical tool in production practice.

\noindent \textbf{Limitation.}
Although the proposed iSegMan achieves flexible, controllable, and interactive 3D scene manipulation, there are a few limitations that need to be addressed.
(\romannumeral1). The semantic editing of 3D Gaussians is limited by the image editor. Although our iSegMan supports the step-by-step achievement of complex editing requirements in an interactive manner, this only alleviates this problem to a certain extent, and each editing step is still limited by the image editor.
(\romannumeral2). The latency of each interaction is limited by the computational cost of the specific manipulation function. For instance, the semantic editing involves gradient-based 3D Gaussian parameter optimization, which restricts the real-time nature of the interaction.
Improving the efficiency of 3D manipulation while maintaining performance is undoubtedly a promising avenue for future exploration.

\noindent \textbf{Acknowledgements.} This work was supported in part by the National Key R\&D Program of China (No. 2022ZD0118201), the Shenzhen Medical Research Funds in China (No. B2302037), National Natural Science Foundation of China (NSFC) under Grant No. 61972217, 32071459, 62176249, 62006133, 62271465, 62406167, and AI for Science (AI4S)-Preferred Program, Peking University Shenzhen Graduate School, China.

\clearpage
\setcounter{page}{1}
\maketitleappendix

\renewcommand{\thefootnote}{\fnsymbol{footnote}}

\renewcommand{\thetable}{\Alph{table}}
\renewcommand{\theequation}{\Alph{equation}}
\renewcommand{\thefigure}{\Alph{figure}}

\setcounter{table}{0}
\setcounter{section}{0}
\setcounter{figure}{0}
\setcounter{equation}{0}

\section{Details of Experimental Settings}

\subsection{Dataset Description}
The datasets used in the experiments are described below:
\begin{itemize}
    \item \textbf{Mip-NeRF 360}~\cite{barron2022mip}. This dataset contains 9 scenes, 5 outdoors and 4 indoors, each of which contains a central object or area with a detailed background.
    \item \textbf{Instruct-N2N}~\cite{haque2023instruct}. This dataset consists of 6 scenes, each of which provides manually captured multi-view natural images, camera poses, and camera paths.
    \item \textbf{LERF}~\cite{kerr2023lerf}. This dataset consists of 9 scenes, each of which provides multi-view images, camera poses, and camera paths.
    \item \textbf{LLFF}~\cite{mildenhall2019local}. This dataset consists of both renderings and real images of natural scenes. The real images are 24 scenes captured by a handheld cellphone.
    \item \textbf{NVOS}~\cite{ren2022neural}. The source data used in this dataset comes from the LLFF dataset, which contains 7 scenes and annotated segmentation masks with 8 instances (two instances are annotated in the ``horn'' scene). This dataset provides a 2D mask ground-truth of the target viewpoints.
    \item \textbf{SPIn-NeRF}~\cite{mirzaei2023spin}. This dataset contains segmentation annotations of 10 scenes, each of which provides 100 multi-view images and corresponding camera poses. For each scene, the first 40 images are the ground-truth captures without the unwanted object, and the rest of the images are the training views with the object present.
\end{itemize}

\subsection{Implementation Details}
All of the original 3D Gaussians in our experiments are trained utilizing the method presented in~\cite{kerbl20233d}, with raw data from publicly available datasets, and rendered during training using the highly optimized renderer proposed in~\cite{kerbl20233d}.
For the epipolar-guided interaction propagation, the default feature extractor for interaction matching employs DINO-small~\cite{caron2021emerging} with a patch size of $16$.\
To improve the efficiency, we perform a $2\times$ downsampling operation on the input image of the feature extractor.
For the visibility-based Gaussian voting, we utilize SAM~\cite{kirillov2023segment} equipped with the ViT-Huge~\cite{dosovitskiy2020image} as the interactive segmenter.
The predetermined threshold of normalized votes is set to $0.8$.
For semantic editing, we employ Instruct-Pix2Pix~\cite{brooks2023instructpix2pix} as the image editor and train each editing instruction for $1500$-$2000$ steps.
We do not apply Gaussian densification during the editing process.
Note that ablation studies are performed on the SPIn-NeRF dataset by default.
We use PyTorch for implementation and a single 32GB NVIDIA V100 GPU for all experiments.

\section{Evaluation Details of Semantic Editing}
\noindent \textbf{User Study.}
The detailed evaluation criteria of user study are presented in~\cref{tab:study}.
We ask the participants to score from three dimensions: accuracy of instruction comprehension, rationality of editing results, and quality of editing results.
The scoring criteria for each dimension are quantified on a scale of $1$ to $5$ inclusive, with no allowance for decimal increments.
Finally, we take the average of the scores of three dimensions as the user study score and provide the 95\% confidence interval.
The user study results reported are the average scores of a total of $30$ participants.

\begin{table*}[t]
\renewcommand{\arraystretch}{1.15}
\begin{tabular*}{\linewidth}{l | c | p{13.5cm}}
\noalign{\hrule height 1.2pt}
\textbf{Dimension} & \textbf{\#Point} & \textbf{Description} \\
\hline
\multirow{5}{*}[-4.7ex]{\textbf{Accuracy}} & 1 & Very poor, the system barely understands the instructions and does not match the user's intention at all. \\
\cline{2-3}
& 2 & Rather poor, the understanding of the instructions is not very accurate, and there are irrelevant areas that are obviously changed. \\
\cline{2-3}
& 3 & Acceptable, the understanding of the instructions is basically correct, and there are basically no irrelevant areas that are obviously changed. \\
\cline{2-3}
& 4 & Fairly good, the understanding of the instructions is relatively accurate, and there are basically no irrelevant areas that have been changed, but there is still room for improvement. \\
\cline{2-3}
& 5 & Very good, the system understands the instructions very accurately and there are no obvious shortcomings. \\
\noalign{\hrule height 0.9pt}
\multirow{5}{*}[-4.7ex]{\textbf{Rationality}} & 1 & Very poor, the result is very unreasonable, there is severe distortion or the original features are completely lost. \\
\cline{2-3}
& 2 & Rather poor, the result is relatively unreasonable, the original features are rarely retained, and irrelevant areas are significantly distorted. \\
\cline{2-3}
& 3 & Acceptable, the result is basically reasonable, the original features are basically identifiable, and the distortion in irrelevant areas is not obvious. \\
\cline{2-3}
& 4 & Fairly good, the result is reasonable, the original features can be accurately identified, and there is a small amount of negligible distortion. \\
\cline{2-3}
& 5 & Very good, the result is clearly reasonable, the original features are fully identifiable and there is no obvious distortion. \\
\noalign{\hrule height 0.9pt}
\multirow{5}{*}{\textbf{Quality}} & 1 & Very poor, texture detail is very blurred, color distribution anomalous. \\
\cline{2-3}
& 2 & Rather poor, texture detail is blurred, color distribution is sometimes anomalous. \\
\cline{2-3}
& 3 & Acceptable, texture detail is slightly blurred, color distribution is basically normal. \\
\cline{2-3}
& 4 & Fairly good, texture detail is relatively clear, color distribution is normal. \\
\cline{2-3}
& 5 & Very good, texture detail is very clear, color distribution is very reasonable. \\
\noalign{\hrule height 1.2pt}
\end{tabular*}
\caption{{The detailed evaluation criteria of the user study.}}
\label{tab:study}
\end{table*}

\noindent \textbf{CLIP Directional Similarity.}
CLIP directional similarity~\cite{gal2022stylegan} refers to the cosine similarity between the change of the images and captions in the CLIP~\cite{radford2021learning} embedding space during the editing process.
CLIP directional similarity measures the consistency of the change between images and captions. The higher the value, the more the edited image matches the editing instructions, and vice versa.
The calculation is presented in~\cref{equ:similarity}.
\begin{equation}
\begin{split}
\label{equ:similarity}
&\Delta \boldsymbol{I} = E_I(\mathcal{I}_v^e) - E_I(\mathcal{I}_v), \\
&\Delta \boldsymbol{T} = E_T(t_e) - E_T(t_{ori}), \\
&\text{CLIP}_{dir} = \frac{\Delta \boldsymbol{I} \cdot \Delta \boldsymbol{T}}{\vert \Delta \boldsymbol{I} \vert \vert \Delta \boldsymbol{T} \vert},
\end{split}
\end{equation}
where $E_I$ and $E_T$ represent the image and text encoders of CLIP, respectively. $\mathcal{I}_v^e$ represents the image rendered from the edited scene, $\mathcal{I}_v$ represents the image rendered from the original scene. $t_e$ represents the caption of the edited image, $t_{ori}$ represents the original image caption. $v$ represents the rendering viewpoint, and we compute the average metric over all viewpoints for each scene.

\section{Results of Ablation Study}

\begin{table}[t]
\centering
\renewcommand{\arraystretch}{1.2}
\setlength{\tabcolsep}{4mm}
\footnotesize
\begin{tabular*}{\linewidth}{c | cccc}
\noalign{\hrule height 1.2pt}
\multirow{2}{*}{\makecell[c]{Epipolar \\ Constraint}} & \multirow{2}{*}{\makecell[c]{mIoU\\(\%)}} & \multirow{2}{*}{\makecell[c]{mAcc\\(\%)}} & \multicolumn{2}{c}{Execution Time} \\
\cline{4-5}
& & & Feature & Segment \\
\hline
\ding{55} & 88.7 & 98.5 & 52s & 7s \\
\rowcolor{aliceblue!80}
\checkmark & \textbf{92.4} & \textbf{99.1} & {52s} & \textbf{6s} \\
\noalign{\hrule height 1.2pt}
\end{tabular*}
\caption{Ablation on epipolar constraint.}
\vspace{-2mm}
\label{tab:constraint}
\end{table}

\noindent \textbf{Epipolar Constraint.}
To verify the effectiveness of the epipolar constraint, we remove it and evaluate the accuracy and execution time of the region selection, \textit{cf.}~\cref{tab:constraint}.
The results show that removing the epipolar constraint does produce incorrectly matched interactions due to the noise introduced by significantly increasing the search space, thus reducing accuracy.

\begin{table}[t]
\centering
\renewcommand{\arraystretch}{1.2}
\setlength{\tabcolsep}{4.73mm}
\footnotesize
\begin{tabular*}{\linewidth}{c | cccc}
\noalign{\hrule height 1.2pt}
\multirow{2}{*}{\makecell[c]{IIM}} & \multirow{2}{*}{\makecell[c]{mIoU\\(\%)}} & \multirow{2}{*}{\makecell[c]{mAcc\\(\%)}} & \multicolumn{2}{c}{Execution Time} \\
\cline{4-5}
& & & Feature & Segment \\
\hline
\ding{55} & 83.9 & 96.4 & 46s & \textbf{5s} \\
\rowcolor{aliceblue!80}
\checkmark & \textbf{90.1} & \textbf{98.2} & 46s & 6s \\
\noalign{\hrule height 1.2pt}
\end{tabular*}
\caption{Ablation on iterative inspection mechanism.}
\vspace{-2mm}
\label{tab:inspection}
\end{table}

\noindent \textbf{Iterative Inspection Mechanism.}
To verify the effectiveness of the iterative inspection mechanism, we also remove it and evaluate the accuracy and execution time, \textit{cf.}~\cref{tab:inspection}.
Since the iterative inspection mechanism only works when the target region is occluded or out of view, we select four scenes with such situations for evaluation, namely ``bicycle'' and ``counter'' from the Mip-NeRF 360~\cite{barron2022mip} dataset, and ``bouquet'' and ``figurines'' from the LERF~\cite{kerr2023lerf} dataset.
We report the average results of four scenes and adopt a uniform sampling rate of 25\% for each scene to  maintain efficiency.
The results indicate that removing the iterative inspection mechanism introduces noise matching interactions that cause incorrect 2D segmentations to participate in the voting, resulting in a decrease in accuracy.

\begin{table}[t]
\centering
\renewcommand{\arraystretch}{1.15}
\setlength{\tabcolsep}{7mm}
\footnotesize
\begin{tabular*}{\linewidth}{c | cc}
\noalign{\hrule height 1.2pt}
Feature Extractor & mIoU(\%) & mAcc(\%) \\
\hline
DINO~\cite{caron2021emerging} & \textbf{92.4} & \textbf{99.1} \\
DINOv2~\cite{oquab2023dinov2} & 92.3 & 99.1 \\
MoCov3~\cite{chen2021empirical} & 92.0 & 98.9 \\
\noalign{\hrule height 1.2pt}
\end{tabular*}
\caption{Ablation on feature extractor.}
\vspace{-2mm}
\label{tab:extractor}
\end{table}

\noindent \textbf{Feature Extractor.}
To test the generalizability of the epipolar-guided interaction propagation, we employ different feature extractors for the ablation, \textit{cf.}~\cref{tab:extractor}.
We employ DINO~\cite{caron2021emerging}, DINOv2~\cite{oquab2023dinov2} and MoCov3~\cite{chen2021empirical} respectively for evaluation, and the results indicate that the proposed method is robust to the feature extractor.

\begin{figure*}[ht]
\hsize=\linewidth
\centering
\includegraphics[width=\linewidth]{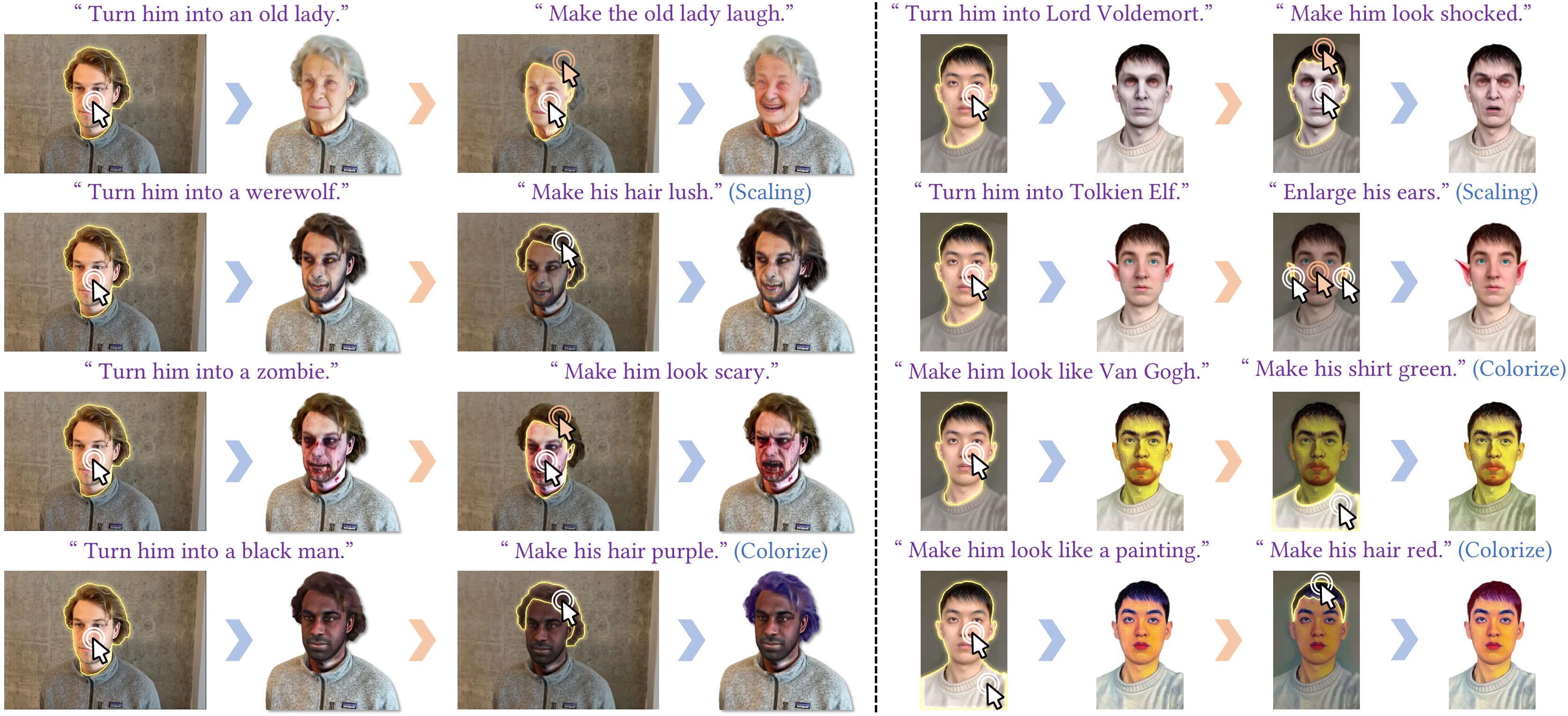}
\vspace{-4mm}
\caption{
\textbf{Additional visualization results.}
Orange arrows indicate interactive 3D segmentation, and blue arrows indicate semantic editing.
}
\vspace{-2mm}
\label{fig:appendix}
\end{figure*}

\section{Preliminary: 3D Gaussian Splatting}

3DGS~(Gaussian Splatting)~\cite{kerbl20233d} models a 3D scene as a set of 3D Gaussian primitives, which are initialized from the sparse point clouds obtained by Structure from Motion~(SfM)~\cite{snavely2006photo}. Each Gaussian $\Theta_i$ is parameterized by a center point $x$ and a covariance matrix $\boldsymbol{\Sigma}_i$, which represents the distribution as:
\begin{equation}
\label{equ:gs}
    \Theta_i (\boldsymbol{x}) = e^{-\frac{1}{2} \textit{$\boldsymbol{x}$}^{T} \boldsymbol{\Sigma}_i^{-1} \textit{$\boldsymbol{x}$} }.
\end{equation}
To derive a physically meaningful covariance matrix that is necessarily positive semi-definite, the subsequent equivalent representation is employed:
\begin{equation}
    \boldsymbol{\Sigma}_i = \boldsymbol{R}_i\boldsymbol{S}_i\boldsymbol{S}_i^{T}\boldsymbol{R}_i^{T},
\end{equation}
where the covariance matrix $\boldsymbol{\Sigma}_i$ is decomposed into a scaling factor $\boldsymbol{S}_i$ and a rotation quaternion $\boldsymbol{R}_i$.
Moreover, an opacity $\sigma_i$ is employed to control the influence of each Gaussian when blending across the scene, and a color $\boldsymbol{c}_i$ is applied to represent its appearance.

To summarize, each 3D Gaussian is parameterized by a set of attributes: position $\boldsymbol{\mu}_i \in \mathbb{R}^3$, scaling factor $\boldsymbol{S}_i \in \mathbb{R}^3$, rotation quaternion $\boldsymbol{R}_i \in \mathbb{R}^4$, opacity $\sigma_i \in \mathbb{R}$, and color $\boldsymbol{c}_i \in \mathbb{R}^k$~(where $k$ indicates the degrees of freedom). 
Each 3D scene can be formally represented by a 3D Gaussian set: $ \Theta = \{ (\boldsymbol{\mu}_i, \boldsymbol{S}_i, \boldsymbol{R}_i, \sigma_i, \boldsymbol{c}_i) \}_{i=1}^N $, where $N$ indicates the number of 3D Gaussians.
These 3D Gaussians can be effectively rendered to compute the color $\boldsymbol{C}$ by blending $N$ ordered Gaussians overlapping the pixel:
\begin{equation}
\label{equ:volume_render}
    \boldsymbol{C} = \sum_{i \in N} \boldsymbol{c}_i \alpha_i \prod_{j=1}^{i-1} (1-\alpha_j),
\end{equation}
where $\alpha_i$ is calculated by evaluating $\Theta_i$ with~\cref{equ:gs} multiplied by its opacity $\sigma_i$.

\section{Additional Visualization Results}
We present additional visualization results, \textit{cf.}~\cref{fig:appendix}.
For semantic editing, we provide text editing instructions, while for other manipulation requirements, we provide requirement descriptions and specify the tools to be invoked (marked in blue).
The extensive and impressive visualization results demonstrate that our iSegMan provides precise region control and excellent manipulation performance, significantly enhancing the controllability, flexibility and practicality of existing 3D manipulation systems.

{
    \small
    \bibliographystyle{ieeenat_fullname}
    \bibliography{main}
}


\end{document}